\documentclass[sigconf]{acmart}

\AtBeginDocument{%
  \providecommand\BibTeX{{%
    \normalfont B\kern-0.5em{\scshape i\kern-0.25em b}\kern-0.8em\TeX}}}

\setcopyright{acmcopyright}
\copyrightyear{2018}
\acmYear{2018}
\acmDOI{10.1145/1122445.1122456}

\acmConference[Woodstock '18]{Woodstock '18: ACM Symposium on Neural
  Gaze Detection}{June 03--05, 2018}{Woodstock, NY}
\acmBooktitle{Woodstock '18: ACM Symposium on Neural Gaze Detection,
  June 03--05, 2018, Woodstock, NY}
\acmPrice{15.00}
\acmISBN{978-1-4503-XXXX-X/18/06}

\usepackage{multirow}
\usepackage{bigstrut}
\usepackage{colortbl}
\usepackage{indentfirst}
\usepackage{color}
\usepackage{makecell}
\usepackage{graphicx}
\usepackage{float}

\usepackage{marvosym}
\usepackage{ifsym}

\acmSubmissionID{icmrs3035}

\newcommand{\eg}{\emph{e.g.,}~}

\newcommand{\ie}{\emph{i.e.,}~}
\newcommand{\wrt}{\emph{w.r.t.}~}
\newcommand{\specialcell}[2][l]{%
  \begin{tabular}[#1]{@{}l@{}}#2\end{tabular}}

\copyrightyear{2022}
\acmYear{2022}
\setcopyright{acmlicensed}\acmConference[ICMR '22]{Proceedings of the 2022 International Conference on Multimedia Retrieval}{June 27--30, 2022}{Newark, NJ, USA}
\acmBooktitle{Proceedings of the 2022 International Conference on Multimedia Retrieval (ICMR '22), June 27--30, 2022, Newark, NJ, USA}
\acmPrice{15.00}
\acmDOI{10.1145/3512527.3531418}
\acmISBN{978-1-4503-9238-9/22/06} 

\begin{document}
\fancyhead{} 
\title{Lesion Localization in OCT by Semi-Supervised Object Detection}


\author{Yue Wu}
\affiliation{%
  \institution{Key Lab of DEKE \\ Renmin University of China}
  }

\author{Yang Zhou}
\affiliation{%
  \institution{Vistel AI Lab \\ Beijing Visionary Intelligence Ltd.}
  }

\author{Jianchun Zhao}
\affiliation{%
  \institution{Vistel AI Lab \\ Beijing Visionary Intelligence Ltd.}
  }

\author{Jingyuan Yang}
\affiliation{%
  \institution{Dept. of Ophthalmology \\ Peking Union Medical College Hospital}
  }

\author{Weihong	Yu}
\affiliation{%
  \institution{Dept. of Ophthalmology \\ Peking Union Medical College Hospital}
  }

\author{Youxin Chen}
\affiliation{%
  \institution{Dept. of Ophthalmology \\ Peking Union Medical College Hospital}
  }

\author{Xirong Li}
\authornote{Corresponding author: Xirong Li (xirong@ruc.edu.cn)}
\affiliation{%
  \institution{Key Lab of DEKE \\ Renmin University of China}
 }


\begin{abstract}
Over 300 million people worldwide are affected by various retinal diseases. By noninvasive Optical Coherence Tomography (OCT) scans, a number of abnormal structural changes in the retina, namely retinal lesions, can be identified. Automated lesion localization in OCT is thus important for detecting retinal diseases at their early stage. To conquer the lack of manual annotation for deep supervised learning, this paper presents a first study on utilizing semi-supervised object detection (SSOD) for lesion localization in OCT images. To that end, we develop a taxonomy to provide a unified and structured viewpoint of the current SSOD methods, and consequently identify key modules in these methods. To evaluate the influence of these modules in the new task, we build OCT-SS, a new dataset consisting of over 1k expert-labeled OCT B-scan images and over 13k unlabeled B-scans. Extensive experiments on OCT-SS identify Unbiased Teacher (UnT) as the best current SSOD method for lesion localization. Moreover, we improve over this strong baseline, with mAP increased from 49.34 to 50.86. 
\end{abstract}



\begin{teaserfigure}
\end{teaserfigure}

\maketitle

\section{Introduction}

\begin{figure}[htb!]
  \centering
  \includegraphics[width=\columnwidth]{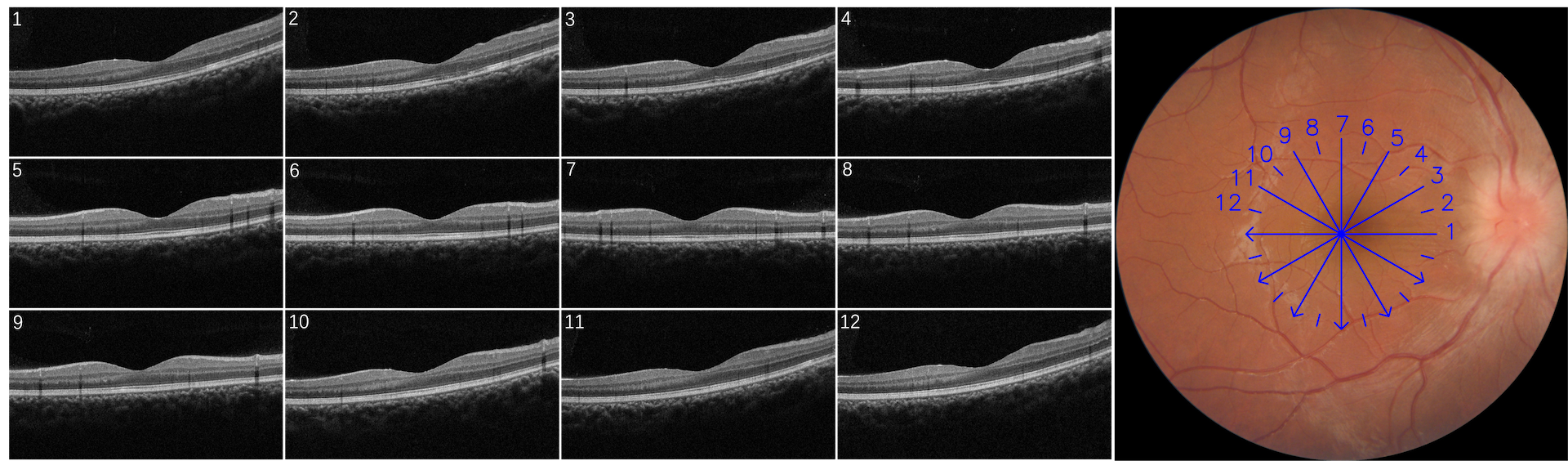}
  \caption{An array of OCT B-scan images and a color fundus photograph (right) acquired during an eye examination. Device: Topcon Maestro-1 (Topcon Corp., Japan). Numbered blue arrows on the color fundus photo are manually added to show the locality of the corresponding OCT images.}
\label{fig:oct-cfp-example}
\end{figure}

Over 300 million people worldwide are affected by various retinal diseases, such as age-related macular degeneration (AMD), diabetic macular edema (DME), and pathologic myopia  \cite{10.1007/978-3-319-66179-7_34,bjo21-mcs}. 
These retinal diseases are among the leading causes of severe vision loss or even blindness, especially in aged populations \cite{LGH}. Luckily, many of the diseases can be identified at their early stage by noninvasive fundus examinations. Optical coherence tomography (OCT) is one of the advanced retinal imaging modalities at present. Compared with color fundus photography, OCT has some unique advantages in identifying abnormal structural changes  in the retina, \emph{a.k.a.} \emph{retinal lesions}. 
OCT provides a cross-sectional analysis of the retina, retinal pigment epithelium, and choroid with depth-resolved segmentation and histology-like resolution \cite{midena2020optical}, see Fig. \ref{fig:oct-cfp-example}. By a volumetric OCT scan, retinal lesions such as pigment epithelial detachment (PED) and intraretinal fluid (IRF) \wrt specific retina diseases can be visualized, see Table \ref{tab:lesion-dis}. 
Lesion localization in OCT images is thus of great social value.

\begin{figure*}[h]
  \centering
  \includegraphics[width=\linewidth]{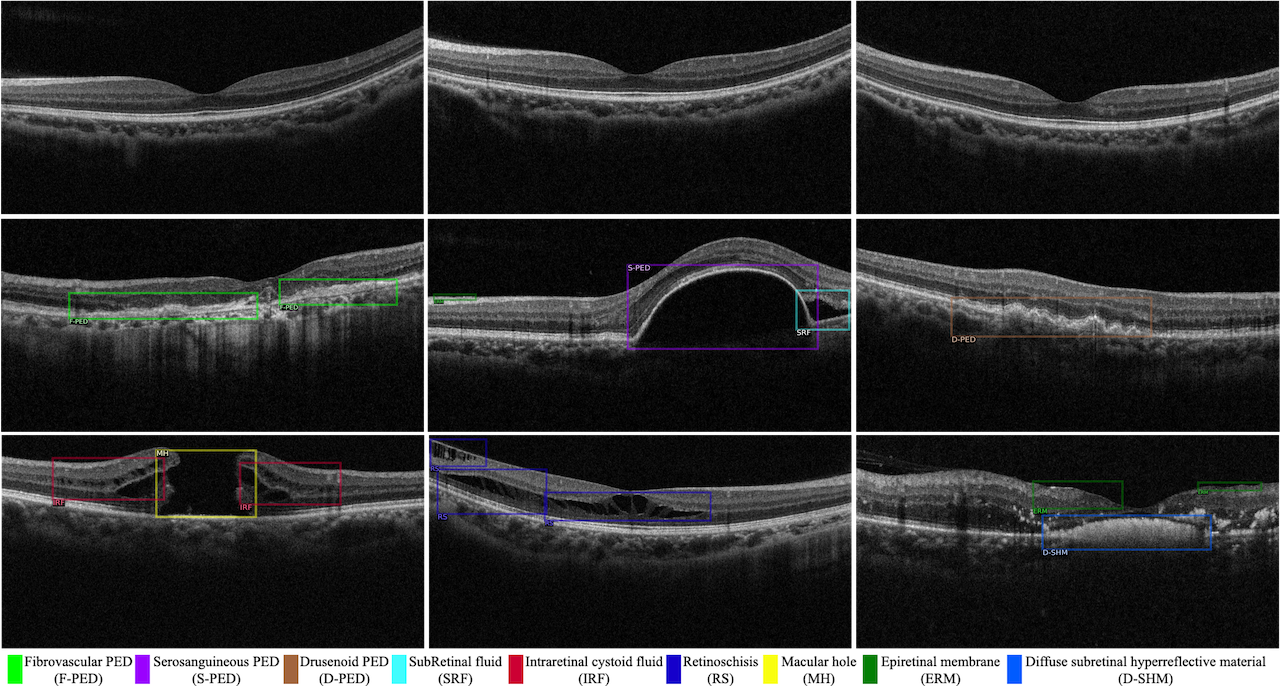}
  \caption{Sample B-scan images of our OCT-SS dataset, built for lesion localization by semi-supervised object detection. The first row is normal B-scans, while the remaining rows are nine retinal lesions labeled by experts and used in this study.}
\label{fig:anno}    
\end{figure*}


Recently, deep learning-based methods, such as image classification, objection detection and semantic segmentation, have been applied successfully on OCT images to solve various  tasks including retinal layer segmentation \cite{10.1007/978-3-319-66179-7_34,10.1007/978-3-030-32239-7_14}, lesion localization \cite{Fan2020PositiveAwareLD} and multi-modal retinal disease recognition \cite{miccai19-amd,LiZWLZDYC21,arxiv-mmcamd}, to name just a few. As these methods follow a standard supervised learning paradigm, a considerable amount of labeled data is often required. However, due to the subtle difference between varied lesions, as exemplified in Fig. \ref{fig:anno}, lesion labeling requires well-trained eyes. Moreover, OCT images arrive in sequences. All this makes region-level manual labeling on the OCT data extremely expensive and time-consuming.


Meanwhile, good progress on semi-supervised object detection (SSOD) has been documented, mostly in the context of natural scene images \cite{sohn2020detection,Zhou_2021_CVPR,liu2021unbiased,xu2021end}. Given a relatively small set of labeled images, the basic idea of SSOD is to exploit a much larger set of unlabeled images to derive an object detector that is better than learning exclusively from the labeled set. Such a setting well suits the task of retinal lesion localization in OCT images.

\begin{table}[tbh!]
   \renewcommand{\arraystretch}{1.2}
    \caption{Nine retinal lesions used in this study and related fundus diseases.}
    \scalebox{0.65}{
    \begin{tabular}{@{}l|l@{}}
    \toprule 
    \textbf{Lesion} & \textbf{Related fundus diseases} \\ \hline
    Fibrovascular PED (F-PED)  &    AMD, Idiopathic choroidal neovascularization      \\ \hline
    Serosanguineous PED (S-PED)  &  AMD, Central serous chorioretinopathy        \\ \hline
    Drusenoid PED (D-PED)  & AMD        \\ \hline
    SubRetinal fluid (SRF)    &  DME, Central serous chorioretinopathy \\ \hline
    Intraretinal cystoid fluid (IRF)    & DME, Central serous chorioretinopathy\\ \hline
    Retinoschisis (RS)     & Retinoschisis\\ \hline
    Macular hole (MH)     & Macular hole      \\ \hline
    Epiretinal membrane (ERM)    & Epiretinal membrane     \\ \hline
   Diffuse subretinal hyperreflective material (D-SHM)  & AMD, Myopic choroidal neovascularization         \\ 
    \bottomrule 
    \end{tabular}
    }
\label{tab:lesion-dis}
\end{table}

Research on SSOD boils down to answering the following two questions, \ie \emph{how to generate pseudo bounding-box labels from unlabeled images?} and \emph{how to exploit such auto-labeled data together with the previously labeled data?} While these questions have been answered with some success, \eg by Instant Teaching (InsT) \cite{Zhou_2021_CVPR}, Unbiased Teacher (UnT)  \cite{liu2021unbiased}, and Soft Teacher (SoftT) \cite{xu2021end}, these efforts are mostly targeted at object detection in natural scenes as provided by the PASCAL VOC and MS-COCO benchmarks. For the natural-scene domain, strong data augmentation operations such as large-scale Cutout or geometric transformation are often performed for better performance \cite{xu2021end}. However, as lesions in an OCT image are closely related to their localities in the image, the original anatomical position information is crucial for lesion localization. Therefore, such information can be disrupted with easy given the aforementioned operations. To what extent can conclusions and good practices of SSOD learned from the natural-image domain be generalized to the OCT-image domain is mostly untouched. 


As the first work on applying SSOD for lesion localization in OCT images, we analyze key modules of the state-of-the-art methods, which consists of pseudo label generation by a teacher network, initialization \& update  of the teacher network, strong data augmentation, and losses for the unlabeled data. The analysis is followed by an extensive experimental comparison. Consequently, we identify a set of good practices for lesion localization in a semi-supervised scenario. To sum up, the main contributions of this paper are:
\begin{itemize}
  \item To the best of our knowledge, we are the first  
  to utilize SSOD techniques for lesion localization in OCT images.  
  \item We develop a taxonomy of the current SSOD methods, see Table \ref{tab:taxonomy}, which provides a unified and structured viewpoint. The taxonomy is important as it allows us to see clearly both common and unique patterns across the existing methods. 
  \item Guided by the taxonomy, we conduct an extensive empirical study to reveal the influence of the key modules in SSOD and consequently sort out the good practices for the new task. Moreover, we develop OCT-SS as a new dataset, which will be released with due ethical approval.
\end{itemize}
For up-to-date information, we refer to our GitHub page (\url{https://github.com/li-xirong/oct-ss}).

\section{Related Work}

\begin{table*}[h]
    \caption{Proposed taxonomy of the state-of-the-art (SOTA) for semi-supervised object detection (SSOD). We identify good and bad practices for OCT lesion localization, which are highlighted in green and red cells, respectively. The integration of the good practices allows us to outperform the best current method (which is UbT) in the new context.
    }
    \scalebox{0.9}{
    \begin{tabular}{@{}l|r|ll|llll|l|ll@{}}
    \toprule
    \multicolumn{1}{c|}{}  &\multicolumn{1}{c|}{}  & \multicolumn{2}{c|}{\textbf{Teacher} $\theta_t$}  & \multicolumn{4}{c|}{\textbf{Strong data augmentation} $A$}  & \multicolumn{1}{c|}{}  & \multicolumn{2}{c}{\textbf{Loss for unlabeled data} $\ell_u$}                                                
    \\ \cline{3-8} \cline{10-11}
    \multirow{-2}{*}{\textbf{Method}} &
    \multicolumn{1}{c|}{\multirow{-2}{*}{\specialcell{5\% COCO \\ mAP }}}&\multicolumn{1}{c|}{\emph{Initialization}} & \multirow{-1}{*}{\emph{Update}} & \multicolumn{1}{c|}{\cellcolor[HTML]{9AFF99}Color} & \multicolumn{1}{c|}{\cellcolor[HTML]{9AFF99}Cutout} & \multicolumn{1}{c|}{\cellcolor[HTML]{FD6864}Mixup} &
    \multicolumn{1}{c|}{\cellcolor[HTML]{FD6864}Geo} &
    \multirow{-2}{*}{\specialcell{\textbf{Pseudo label generation}\\(PLG)}} & \multicolumn{1}{c|}{$\ell_{u,cls}$} & \multicolumn{1}{c}{$\ell_{u,reg}$}                 
    \\ 
    \hline
    STAC \cite{sohn2020detection} & \multicolumn{1}{c|}{24.38} & \multicolumn{1}{l|}{\cellcolor[HTML]{9AFF99}Burn-In} & -- & \multicolumn{1}{l|}{\checkmark} & \multicolumn{1}{l|}{\checkmark} & \multicolumn{1}{l|}{--} &\multicolumn{1}{l|}{\checkmark} &
    \multicolumn{1}{c|}{Score filter} & \multicolumn{1}{l|}{\cellcolor[HTML]{FFFFFF}CE}  & \multicolumn{1}{c}{L1} 
    \\ \hline
    InsT \cite{Zhou_2021_CVPR} &  \multicolumn{1}{c|}{26.75} & \multicolumn{1}{l|}{ImageNet} & BP & \multicolumn{1}{l|}{\checkmark} & \multicolumn{1}{l|}{\checkmark} & \multicolumn{1}{l|}{\checkmark}& \multicolumn{1}{l|}{--}& \multicolumn{1}{c|}{Score filter} & \multicolumn{1}{l|}{CE} &  \multicolumn{1}{c}{L1} 
    \\ \hline
    UbT \cite{liu2021unbiased} & \multicolumn{1}{c|}{28.27} & \multicolumn{1}{l|}{Burn-In} & \cellcolor[HTML]{9AFF99}EMA & \multicolumn{1}{l|}{\checkmark} &\multicolumn{1}{l|}{\checkmark} & \multicolumn{1}{l|}{--}& \multicolumn{1}{l|}{--}& \multicolumn{1}{c|}{Score filter}
    &\multicolumn{1}{l|}{\cellcolor[HTML]{9AFF99}FL} &  \multicolumn{1}{c}{--}                  
    \\ \hline
    SoftT \cite{xu2021end} &\multicolumn{1}{c|}{30.74} &\multicolumn{1}{l|}{ImageNet} & EMA & \multicolumn{1}{l|}{\checkmark} & \multicolumn{1}{l|}{\checkmark} & \multicolumn{1}{l|}{--}& \multicolumn{1}{l|}{\checkmark}& \multicolumn{1}{c|}{\cellcolor[HTML]{9AFF99}Double filter} & \multicolumn{1}{l|}{WCE} & \multicolumn{1}{c}{L1} 
    \\ \hline
    UbT+ (\emph{this paper}) &\multicolumn{1}{c|}{--} &\multicolumn{1}{l|}{Burn-In} & EMA & \multicolumn{1}{l|}{\checkmark} & \multicolumn{1}{l|}{\checkmark} & \multicolumn{1}{l|}{--}& \multicolumn{1}{l|}{--}& \multicolumn{1}{c|}{Double filter} & \multicolumn{1}{l|}{FL} & \multicolumn{1}{c}{L1} 
    \\ 
    \bottomrule
    \end{tabular}
    \label{tab:taxonomy}    
    }
\end{table*}

{\bfseries Semi-supervised object detection.}
The SSOD setting we discuss is that the training contains a small set of labeled data and another set of completely unlabeled data (i.e., only images). In this setting, CSD\cite{NEURIPS2019_d0f4dae8} proposed a method based on consistency regularization, which enforces the predictions of an input image and its flipped version to be consistent. However, in the early phase of training, consistency regularization  regularizes the model towards high entropy predictions, and prevents it from achieving good accuracy\cite{sohn2020fixmatch}. Recently, these works STAC\cite{sohn2020detection}, Instant Teaching\cite{Zhou_2021_CVPR}, Unbiased Teache\cite{liu2021unbiased}, Soft Teacher\cite{xu2021end}, which use pseudo labeling with weak-strong data augmentation scheme for model training, are the current state-of-the-arts. 


Apparently, the upper limit of performance of the framework depends on the quality of the pseudo label, as we rely on the pseudo label to train the student model. There are two ways to improve the quality of pseudo label. 1) Generate better quality box candidates. 2) Filter out box candidates of poor quality. Therefore a good initialization \& update strategy for teacher model and a good pseudo label generation strategy are very important. 

The weak-strong data augmentation scheme is actually augmentation driven consistency regularization, which enforces the model to maintain consistent predictions between the weakly augmented and the strongly augmented unlabeled data, and thus encourages the model to learn useful information from the pseudo annotations. And the teacher model use unlabeled data applied weak augmentations is due to the consideration of the accurate of pseudo label. The strength of augmentations is a relative concept and has not been clearly defined. Intuitively, the key of weak-strong data augmentation scheme lies in the difference between weak augmentations and strong augmentations. When the weak augmentations remain unchanged, the more complex and appropriate the strong augmentations, the more information the model can learn from the pseudo label\cite{Zhou_2021_CVPR}.

Table \ref{tab:taxonomy} systematically shows the different strategies for design points of the current SOTA of SSOD, which consists of strong data augmentations, initialization \& update strategy of teacher network, pseudo label generation strategy, unsupervised loss functions. The green and red colors in the table indicate that in our experimental evaluation, the corresponding strategy has positive and negative effects on OCT lesion localization, respectively. In section \ref{section:method}, we interpret these strategies in detail.

{\bfseries Semi-supervised lesion localization.}
FocalMix \cite{Wang_2020_CVPR} is the first to investigate the problem of semi-supervised learning for lesion localization. Their task is lung nodule localization in thoracic CT images. FocalMix propose a pseudo label generation strategy that leverages anchor-level ensembles of augmented image patches by rotation and flipping, adapt the Focal-loss for soft-target and adapt the MixUp augmentation at both the image level and object level in light of unique characteristics of the lung nodule localization. Compared with lung nodule localization(single category lesion localization), OCT lesion localization is more challenging for the OCT lesions have more categories and more complex morphologies. Some lesions have similar features, such as intraretinal cystoid fluid (IRF) and retinoschisis (RS) shown in Figure \ref{fig:anno}. 

{\bfseries OCT lesion localization.}
Yang et al.\cite{9098380} proposes an unsupervised domain adaptation framework for cross-device OCT lesion localization via learning adaptive features. They integrate global and local adversarial discriminators into Faster R-CNN\cite{ren15fasterrcnn}, and apply L2-norm function to the global feature to stabilize the discrimination in target domain. They select subRetinal fluid (SRF), choroidal neovascularization (CNV), and retinal pigment epithelium atrophy (RPEA) as detecting lesions. Wang et al.\cite{10.1167/tvst.9.2.46} developed an intelligent system using Feature pyramid networks (FPN)\cite{Lin_2017_CVPR} for OCT lesion localization and making urgent referrals through a decision network that used both localization results and thickness maps as input. Fan et al.\cite{Fan2020PositiveAwareLD} proposes a positive-aware lesion localization network with cross-scale feature pyramid(CFP) based on Faster R-CNN to detect 9 categories of retinal lesions on OCT images. Different from FPN, the features at different scales of CFP combine the information of all other scales. They generate positive-aware lesion confidence at global level and regional level. And the activation response on global feature map and positive proposal confidence are integrated into the final localization score. Prior works\cite{10.1167/tvst.9.2.46},\cite{Fan2020PositiveAwareLD} have already demonstrated promising results in various OCT lesion localization, but the success should be attributed to not only recent progress in deep learning techniques but also large volumes of carefully labeled data. Note that the data set they used contains nearly 40,000 B-scans with lesion labels.

\section{Taxonomy of SSOD Methods}
\label{section:method}

\subsection{Problem Formalization}

Given a set of $n_l$ labeled images $D_{l}=\{(x_{l},y_{l})\}$ and a much larger set of $n_u$ \emph{unlabeled} images $D_u=\{x_u\}$, the goal of SSOD is  can be stated as follows. By jointly exploiting both $D_l$ and $D_u$, one aims to obtain an object detector which is better than its counterpart trained on $D_l$ alone. 

Given an object detection network parameterized by $\theta$, we use $p(c,b| x, \theta)$ to indicate the network-predicted probability of having an object of class $c$ at region $b$ in a specific image $x$. Naturally, in order to optimize $\theta$, the label $y_l$ shall consist of a set of manually labeled bounding-box (bbox) annotations $\{(c_l, b_l)\}$. In a similar vein, we use $\{(c_u, b_u)\}$ to denote a set of \emph{pseudo} labels to be extracted from $x_u$. By definition, such pseudo labels are meant for guiding the training process of a targeted object detection network. Hence, following \cite{liu2021unbiased,xu2021end}, we term the model producing the pseudo labels a teacher network, parameterized by $\theta_t$. Accordingly, we call the targeted network a student parameterized by $\theta_s$. To simplicity our notation, we shorten $p(c,b|x,\theta_s)$ as $p_s(c,b|x)$ and $p(c,b|x,\theta_t)$ as $p_t(c,b|x)$.

\subsection{A Unified Framework}

By analyzing the working pipelines of the current SSOD methods \cite{sohn2020detection,Zhou_2021_CVPR,liu2021unbiased,xu2021end}, we see common patterns and consequently reach a unified description of these methods as follows:
\begin{enumerate}
    \item Generate pseudo labels $\{(c_u,b_u)\}$ for an unlabeled image $x_u$, typically by a heuristic Pseudo Label Generation (PLG) function which takes $p_t(c,b|x_u)$ as input. 
    \item Compute a combined loss $\ell_s + \lambda_u \ell_u$, where $\ell_s$ is a standard object detection loss computed on an labeled image $x_l$ using $p_s(c,b|x_l)$ and $y_l$, while $\ell_u$ is the counterpart of $\ell_s$ computed on the unlabeled image using $p_s(c,b|x_u)$ and $\{(c_u,b_u)\}$. The hyper-parameter $\lambda_u$ balances the two losses.
    \item Update the student network $\theta_s$ by SGD \wrt the combined loss.
    \item Optionally update the teacher network  $\theta_t$ with $\theta_s$. 
\end{enumerate}

The above four-step procedure is executed in an iterative manner until certain stop criterion is met, say reaching a predefined maximal number of iterations. The unified description allows us to develop a taxonomy, see Table \ref{tab:taxonomy}, that provides a structured overview of the current SSOD methods.

\subsubsection{Strategies for Pseudo Label Generation} 

As shown in Table \ref{tab:taxonomy}, two strategies exist for pseudo label generation, \ie score filter and double filter. Given a specific bbox $b$ predicted by the teacher network, the score filter accepts $b$ if its probability of being an foreground object exceeds a pre-specified threshold $\tau$. The foreground score is defined as the maximum probability of all non-background classes. The score filter has been used in STAC \cite{sohn2020detection}, InsT \cite{Zhou_2021_CVPR} and UbT \cite{liu2021unbiased}. Note that the foreground score does not directly measure the quality of the bbox. Probably due to this concern, UbT considers only the pseudo class $c_u$, with the pseudo bbox $b_u$ discarded,  when computing $\ell_u$.

For better bbox selection, the double filter is recently introduced by SoftT \cite{xu2021end}. Given a bbox $b$ preserved after the score filter, a box regression variance $r(b)$ is estimated to measure the reliability of the bbox. Accordingly, $b$ will be rejected if $r(b)$ exceeds a predefined threshold $\tau_2$.


\begin{figure*}[thb]
  \centering
  \includegraphics[width=\linewidth]{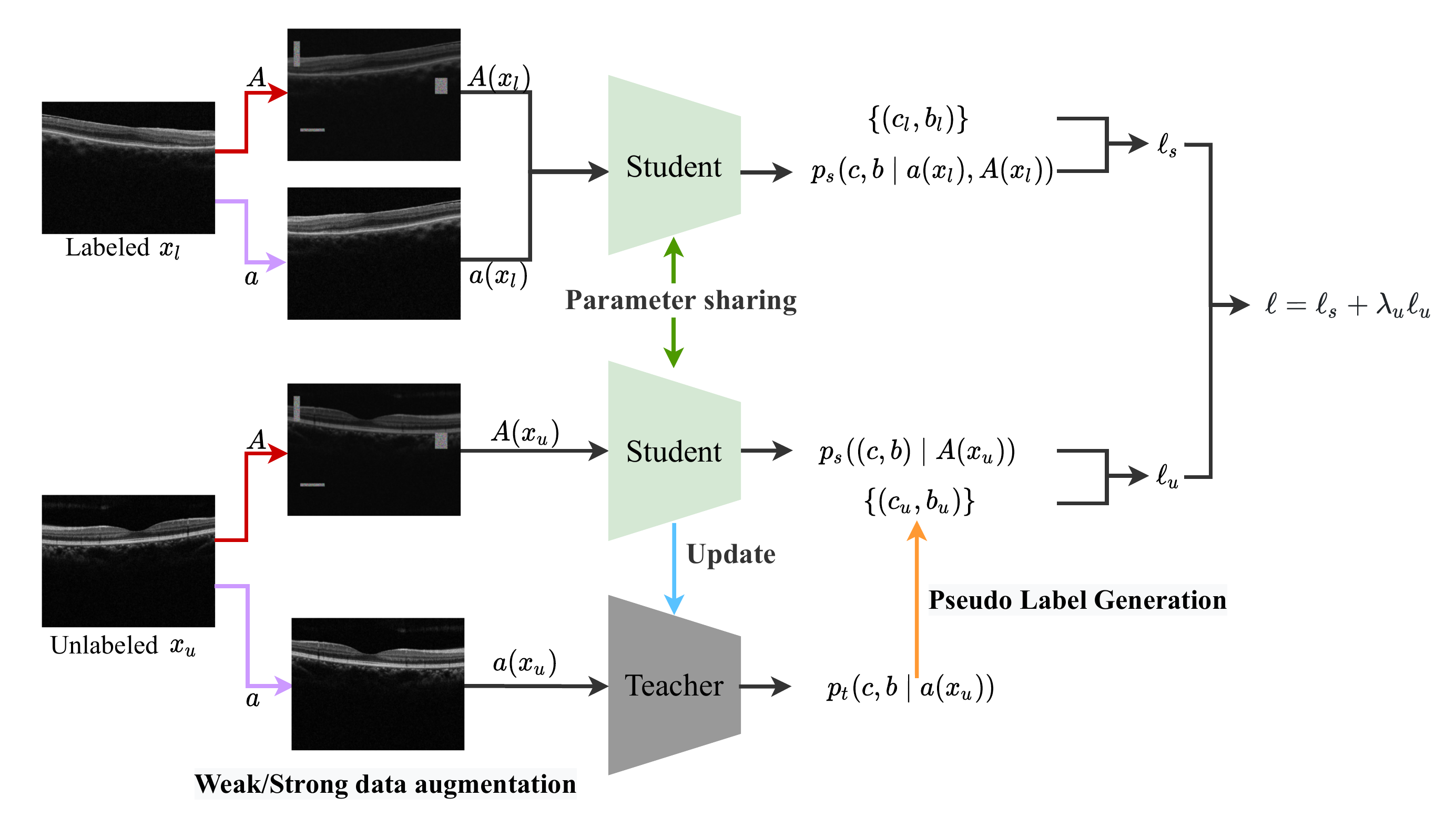}
  \caption{A unified illustration of semi-supervised lesion detection in the new context of retinal lesion localization in OCT images. With the assistance of a teacher network, labeled and unlabeled images are jointly exploited to train a better student network. The student network will be used in the inference stage.  
  }
\label{fig:framework}    
\end{figure*}



\subsubsection{Strong Data Augmentation Operations}
In order to diversify the training data, certain strong data augmentation operation, denoted by $A$, is often performed on both labeled and unlabeled images. Commonly used operations consist of the following three types, \ie  1) \emph{color based} including brightness / contrast / sharpness jitter, 2) \emph{geometric} including x-y  translation, rotation, and shearing, and lastly 3) \emph{bbox manipulation} including Mixup and Cutout. A student network's prediction \wrt a strongly-augmented image is denoted by $p_s(c,b|A(x))$, and $p_s(c,b|a(x))$ for its weakly-augmented counterpart. In a similar manner, we shall have $p_t(c,b|A(x))$, and $p_t(c,b|a(x))$ for the teacher network.

\subsubsection{Initialization and Update Strategies for the Teacher Model}
A well initialized and adaptively updated teacher model can generate better pseudo labels. InsT and SoftT choose to initialize their teacher with a ImageNet-pretrained model. By contrast, STAC and UbT opt for Burn-in, which uses the labeled image set $D_l$ to train the teacher. Burn-in is reported to be beneficial for improving the convergence speed of the student model. As for the update strategy, earlier work either does not update the teacher, as done in STAC, or updates by back propagation (BP), as done in InsT. More recent models (UbT and SoftT) use Exponential Moving Average (EMA) to let the teacher update and improve slowly.



\subsubsection{Loss for unlabeled data}

The loss for unlabeled data $\ell_u$ determines how the pseudo labels $\{(c_u,b_u)\}$ are actually exploited. At a high level, $\ell_u$ can be expressed by $\ell_{u,cls} + \beta \cdot \ell_{u,reg}$, where $\ell_{u,cls}$ is a classification loss measuring the divergence between $p_s(c,b|x_u)$ and $c_u$, while $\ell_{u,reg}$ is a bbox regression loss measuring the spatial gap between the predicted bbox and $b_u$. The hyper-parameter $\beta$ is to balance the two sub losses. Choices for $\ell_{u,cls}$ vary, including Cross-Entropy (CE), Weighted CE (WCE), and Focal Loss (FL), as summarized in  Table \ref{tab:taxonomy}. By contrast, for $\ell_{u,reg}$, all methods consistently use the generalized L1 loss \cite{ren15fasterrcnn}, except for UbT which computes only $\ell_{u,cls}$ with $\beta=0$.


A unified illustration of how the aforementioned modules interact within an SSOD framework is given in Fig. \ref{fig:framework}.

\section{Data Acquisition}

To investigate OCT lesion localization in a semi-supervised setting, we built a real-world dataset termed OCT-SS as follows. OCT-SS will be released with due ethical approval. 

\textbf{OCT Image Acquisition}. We collected OCT data at the outpatient clinic, the Department of Ophthalmology in a state hospital from July 2020 to January 2021. Per eye examination, a Topcon Maestro-1 (Topcon Corp., Japan) multi-modal fundus camera was used in a radial scan mode to simultaneously capture an array of 12 OCT B-scan images and a color fundus photo, see Fig. \ref{fig:oct-cfp-example}. In total, over 1,400 exam cases were gathered from 920 patients (526 females, 394 males, aging between 6 and 91) as a candidate dataset.

\textbf{Lesion Labeling by Experts}. 
An expert committee composed of four retinal specialists was formed and to label a random subset according to dozens of pre-specified lesions related to varied retinal diseases. For quality control, each case was labeled by two experts independently. In case of disagreement, a third expert was asked to make the final decision. Considering the potential spatial correlation between B-scans of a specific case,  each annotator was asked to choose one or two B-scans from a case, and label lesion regions accordingly. 
Note that not all lesions had a reasonable amount of B-scans for training and evaluation. With rarely occurred lesions excluded, we compile a list of nine lesions, see Table \ref{tab:lesion-dis}. A set of 1,085 B-scan images from 541 patients are labeled. Some manual annotations are visualized in Fig. \ref{fig:anno}.



\textbf{Data division}. We divide the labeled dataset at random into training and test, on the basis of patient identities. As such, B-scans from a specific patient appear exclusively in the training set or in the test set, not both. Unlabeled data (with patients in the test set excluded) are used as $D_u$. Table \ref{tab:data} summarizes basis statistics of our experimental data. 

\begin{table}[!ht]
    \renewcommand{\arraystretch}{1.1}
    \caption{Statistics of the OCT-SS dataset from this paper.}
    \scalebox{0.9}{
    \begin{tabular}{@{}l|r|r|rrr@{}}
    \toprule 
    \multirow{2}{*}{\textbf{Data split}} &\multicolumn{1}{c|}{\multirow{2}{*}{\textbf{Patients}}}&
    \multicolumn{1}{c|}{\multirow{2}{*}{\textbf{B-scans}}}& \multicolumn{3}{c}{\textbf{Lesions per B-scan}} 
    \\ \cline{4-6} 
     & & & \textit{min} & \textit{max} & \textit{mean}
    \\ \hline
    Labeled training data $D_l$ &308 &  607 & 1 &  11 &  2.2
    \\  \hline
    Unlabeled training data $D_u$ & 687 & 13,541 &  / &  / & /
    \\ \hline
    Labeled test data & 233 & 478 & 1  & 46  & 2.5
    \\
    \bottomrule 
    \end{tabular}
 }
    \label{tab:data}    
\end{table}

\section{Evaluation}
\begin{table*}[thb!]
    \renewcommand{\arraystretch}{1.1}
    \caption{Performance of varied (SSOD) methods for retinal lesion localization in OCT images. Compared to the supervised  baseline which is exclusively trained on labeled data, the SSOD methods provide better performance. Our proposed UbT+ further improves over the best baseline, \ie UbT, with mAP increasing from 49.34 to 50.86. }
    \scalebox{0.73}{
    \begin{tabular}{@{}llllllcrrrrrrrrrrr@{}}
    \toprule
    \multicolumn{1}{@{}l|}{\multirow{2}{*}{\textbf{Method}}} & \multicolumn{4}{c|}{\textbf{Strong data augmentation}} & \multicolumn{1}{c|}{\multirow{2}{*}{\textbf{PLG}}} & \multicolumn{2}{l|}{\textbf{Loss of unlabeled data}}  & \multirow{2}{*}{mAP} & \multirow{2}{*}{F-PED} & \multirow{2}{*}{S-PED} & \multirow{2}{*}{D-PED} & \multirow{2}{*}{SRF} & \multirow{2}{*}{IRF} & \multirow{2}{*}{RS} & \multirow{2}{*}{MH} & \multirow{2}{*}{ERM} & \multirow{2}{*}{D-SHM} 
    \\ \cline{2-5} \cline{7-8}
    \multicolumn{1}{c|}{} & \multicolumn{1}{c|}{Color} & \multicolumn{1}{l|}{Cutout}                    & \multicolumn{1}{l|}{Mixup}  & \multicolumn{1}{l|}{Geo}   & \multicolumn{1}{l|}{} & \multicolumn{1}{l|}{$\ell_{u,cls}$} & \multicolumn{1}{c|}{$\ell_{u,reg}$} &  &  &   &    &    &  &     &   &  &   
    \\ \hline
    \multicolumn{18}{@{}l}{\emph{Baselines:} }                         
    \\ \hline
    Supervised \cite{ren15fasterrcnn} & \multicolumn{1}{c|}{--} & \multicolumn{1}{l|}{--} & \multicolumn{1}{l|}{--} & \multicolumn{1}{l|}{--} & \multicolumn{1}{l|}{--} & \multicolumn{1}{l|}{--} & \multicolumn{1}{c|}{--} & 43.94 &	24.19 &	32.45 &	33.03 	&54.23 & 53.27 &45.50 &	68.77 &	57.55 &	26.44
     \\ \hline
    Supervised  & \multicolumn{1}{c|}{\checkmark} & \multicolumn{1}{l|}{\checkmark} & \multicolumn{1}{l|}{--} & \multicolumn{1}{l|}{--} & \multicolumn{1}{l|}{--} & \multicolumn{1}{l|}{--} & \multicolumn{1}{c|}{--} & 45.06 &	25.29 &	37.20 & 37.22 & 53.89 & 54.23 &34.75 &72.28 &	58.64 &	31.79
    \\ \hline
   SoftT \cite{xu2021end} & \multicolumn{1}{c|}{\checkmark} & \multicolumn{1}{l|}{\checkmark} & \multicolumn{1}{l|}{--} & \multicolumn{1}{l|}{\checkmark} & \multicolumn{1}{l|}{Double filter}  & \multicolumn{1}{l|}{WCE} & \multicolumn{1}{c|}{L1} & 46.48 &	21.59&	\textbf{52.99} & 41.56&	53.93&	53.96&	35.64&	69.10&	48.51&	40.98
    \\ \hline
    UbT \cite{liu2021unbiased} & \multicolumn{1}{c|}{\checkmark} & \multicolumn{1}{l|}{\checkmark} & \multicolumn{1}{l|}{--} & \multicolumn{1}{l|}{--} & \multicolumn{1}{l|}{Score filter}  & \multicolumn{1}{l|}{FL} & \multicolumn{1}{c|}{--} & 49.34 & 29.87 &	43.86 &	43.73 &	58.06 &	55.85 &	42.65 & \textbf{80.22} & 58.17 & 31.63
    \\ \hline
    \multicolumn{18}{@{}l}{\emph{Variants evaluated by this paper:}}  
    \\ \hline
    UbT (Cutout$\rightarrow$Mixup) & \multicolumn{1}{c|}{\checkmark} & \multicolumn{1}{l|}{--} & \multicolumn{1}{l|}{\checkmark} & \multicolumn{1}{l|}{--} & \multicolumn{1}{l|}{Score filter} & \multicolumn{1}{l|}{FL} & \multicolumn{1}{c|}{--} & 45.65 &	29.52 &	41.41 &	35.56 &	57.88 &	57.13 &	33.98 &	63.80  &\textbf{59.23} &32.33     
    \\ \hline
    UbT (FL$\rightarrow$WCE) & \multicolumn{1}{c|}{\checkmark} & \multicolumn{1}{l|}{\checkmark} & \multicolumn{1}{l|}{--} & \multicolumn{1}{l|}{--} & \multicolumn{1}{l|}{Score filter}  & \multicolumn{1}{l|}{WCE} & \multicolumn{1}{c|}{--} & 47.73&	29.31& 	47.28&	38.47& 	51.52& 	53.12& 	45.12&	71.99& 	54.64& 	36.64
    \\ \hline
   UbT w/ $\ell_{u,reg}$ & \multicolumn{1}{c|}{\checkmark} & \multicolumn{1}{l|}{\checkmark} & \multicolumn{1}{l|}{--} & \multicolumn{1}{l|}{--} & \multicolumn{1}{l|}{Score filter} & \multicolumn{1}{l|}{FL} & \multicolumn{1}{c|}{L1} & 47.83& 31.62& 	39.49&	\textbf{43.83} & 	55.66& 	55.50& 	38.44& 	74.29&	56.59& 	35.02         
    \\ \hline
    UbT w/o Cutout & \multicolumn{1}{c|}{\checkmark} & \multicolumn{1}{l|}{--} & \multicolumn{1}{l|}{--}  & \multicolumn{1}{l|}{--} & \multicolumn{1}{l|}{Score filter} & \multicolumn{1}{l|}{FL} & \multicolumn{1}{c|}{--}  & 48.22 & 30.77 & 42.78 & 41.57 & 58.27& 55.68 & 45.58 & 73.22 & 56.85 &29.27 
    \\ \hline
    UbT (FL$\rightarrow$CE) & \multicolumn{1}{c|}{\checkmark} & \multicolumn{1}{l|}{\checkmark} & \multicolumn{1}{l|}{--} & \multicolumn{1}{l|}{--} & \multicolumn{1}{l|}{Score filter} & \multicolumn{1}{l|}{CE} & \multicolumn{1}{c|}{--} & 48.55& \textbf{38.98}& 41.49& 39.68& 58.00& 	57.71& 	37.90& 	75.34& 	53.53& 	34.32 
    \\ \hline
    SoftT w/o Geo & \multicolumn{1}{c|}{\checkmark} & \multicolumn{1}{l|}{\checkmark} & \multicolumn{1}{l|}{--} & \multicolumn{1}{l|}{--} & \multicolumn{1}{l|}{Double filter} & \multicolumn{1}{l|}{WCE} & \multicolumn{1}{c|}{L1}  & 48.73 &32.92&	51.39&	41.66&	\textbf{59.51} &	56.24&	36.61&	69.10&	51.38&	39.74
    \\ \hline
    UbT+ (\emph{this work})   & \multicolumn{1}{c|}{\checkmark} & \multicolumn{1}{l|}{\checkmark} & \multicolumn{1}{l|}{--} & \multicolumn{1}{l|}{--}   & \multicolumn{1}{l|}{Double filter}        & \multicolumn{1}{l|}{FL} & \multicolumn{1}{c|}{L1}  & \textbf{50.86} & 33.61 &	47.33 &	40.83 &	55.69 &	\textbf{58.26} & \textbf{49.92} &	75.39 	&54.14 &	\textbf{42.52}
    \\
    \bottomrule
    \end{tabular}
    \label{tab:eval}    
}
\end{table*}

\subsection{Experimental Setup}

{\bfseries Baselines}. We compare with UbT \cite{liu2021unbiased} and SoftT \cite{xu2021end}, two leading methods for SSOD, as demonstrated by their superior performance on the MS-COCO benchmark, see Table \ref{tab:taxonomy}. In addition, we include a Faster R-CNN \cite{ren15fasterrcnn}  trained on the labeled data as a supervised baseline. We follow the hyper-parameter settings as used in the original papers, unless otherwise stated.

{\bfseries Details of implementation}.
For all models evaluated in this study, we use the following common setup. For both student and teacher networks, we use Faster RCNN with FPN \cite{Lin_2017_CVPR} and ResNet-50 \cite{He_2016_CVPR} as its backbone. Anchors with 5 scales and 3 aspect ratios are used. We use SGD as the optimizer, with momentum of 0.9 and weight decay of 0.0001 and a batch size of 8. The initial learning rate is 0.01, with a warm-up strategy to adjust the learning rate. 
Per setup, we repeat the training procedure three times and report the best result. As large-scale Cutout may completely cut a lesion out, we reduce the scale and ratio parameters of the Cutout operation. 
All experiments are run with PyTorch. 

{\bfseries Performance metric}.  
We use mean Average Precision (mAP). The IoU threshold is set to $0.3$ as overlap between the prediction and the ground truth at this level is clinically sufficient.


\subsection{Experiment 1. Comparison between SOTA}

To investigate to what extent conclusions drawn on the COCO benchmark can be generalized to the OCT domain, we first make a comparison between the two SOTA methods, UbT and SoftT. Their performance, together with the supervised baselines, is reported in Table \ref{tab:eval}. 
Compared to the best supervised baseline, which has an mAP of 45.06, UbT and SoftT obtain mAP of 49.34 and 46.48, respectively. Both methods are better than the baseline. However, in contrast to their performance on COCO, UbT is superior to SoftT for the new task. 

Note that the two methods differ in multiple aspects including strong data augmentation, pseudo label generation, and $\ell_u$. In what follows, we conduct a series of ablation study to reveal the influence of the individual components. 



\subsection{Experiment 2. Evaluating Strong Data Augmentation Strategies}


With Cutout replaced by Mixup, mAP of UbT drops from 49.34 to 45.65, see the row of Table \ref{tab:eval} starting with UbT (Cutout$\rightarrow$ Mixup). Meanwhile, simply removing Cutout also leads to a drop from 49.34 to 48.22, see the row of UbT w/o Cutout. The result suggests that Cutout has a positive effect on the performance, while the effect of Mixup is negative. 

With the geo operation removed, mAP of SoftT increases from 46.48 to 48.73, see the row of Soft w/o Geo. 
The observation verifies our hypothesis that geometric transformation and Mixup will affect the original anatomical position information of fundus, which is  important for OCT lesion identification. 
In sum, color plus small-scale Cutout are preferred. 


\subsection{Experiment 3. Evaluating PLG Strategies} \label{ssec:exp-plg}
The effectiveness of the SSOD framework depends on the quality of the pseudo labels. Given pseudo labels produced by the score filter, we tried to include the bbox regression loss $\ell_{u,reg}$ into UbT. As the row of UbT (w/ $\ell_{u,reg}$) shows, the performance drops from 49.34 to 47.83. The result suggests that using the score filter alone is insufficient. 

So in what follows, we borrow the double filter from SoftT. As the last row shows, the combination of UbT and the double filter boosts the lesion localization performance, increasing mAP from 49.34 to 50.86. The double filter is found to be useful.


\subsection{Experiment 4. Loss for Unlabeled Data}

Given UbT as the SSOD method, the preferred order of the choices for the classification loss $\ell_{u,cls}$ is the following: FL (49.34), CE (48.55), and WCE (47.73). As for the bbox regression loss $\ell_{u,reg}$, it shall be used together with the double filter, as noted in Section \ref{ssec:exp-plg}. Hence, for the optimal performance, UbT shall be modified, with its score filter replaced by the double filter and with $\ell_{u,reg}$ included. We term the enhanced variant of UbT as UbT+. 



\begin{figure*}[tbh!]
  \centering
  \includegraphics[width=\linewidth]{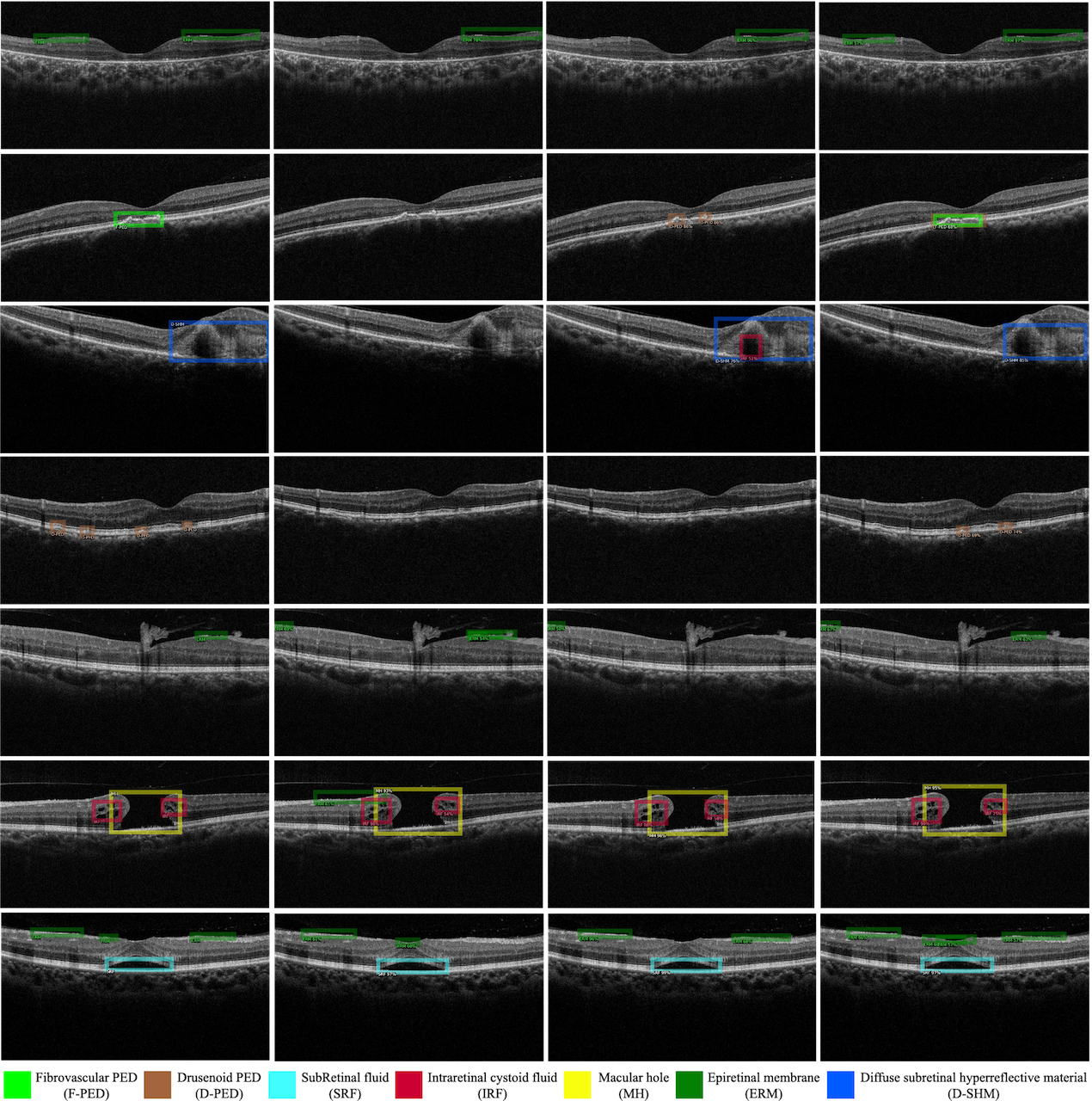}
  \caption{Lesion localization results, read from left to right: ground truth, the supervised baseline, UbT and our UbT+.}
\label{fig4}    
\end{figure*}

\section{Conclusions}
As the first work on applying SSOD for lesion localization in OCT images, we analyze key modules of the state-of-the-art methods, which consists of pseudo label generation by a teacher network, initialization \& update of the teacher network, strong data augmentation, and losses for the unlabeled data. The analysis is followed by an extensive experimental comparison.  Consequently, we identify a set of good practices for lesion localization in a semi-supervised scenario. Specifically, we get the strong data augmentation strategy--Color + small-scale Cutout, which are suitable for OCT images. And we observed that geometric transformation and Mixup impedes the effectiveness of weak-strong data augmentation scheme for affecting the original anatomical position information of fundus, which is very important for OCT lesion identification. And the double filter is found to be a useful pseudo label generation strategy. Hence, for the optimal performance, UbT shall be modified, with its score filter replaced by the double filter and with $\ell_{u,reg}$ included. We term the enhanced variant UbT+, which achieves the best performance on the task of semi-supervised lesion localization in OCT images.

Our current model performs training and inference at the image level, without considering the volumetric information. As adjacent images in a given OCT scan are spatially connected, we shall exploit such information in our future work.


\medskip 
\textbf{Acknowledgments}. This research was supported by NSFC (No. 62172420), BJNSF (4202033), BJNSF-Haidian Original Innovation Joint Fund (19L2062), the Fundamental Research Funds for the Central Universities and the Research Funds of Renmin University of China (No. 18XNLG19), and Public Computing Cloud, Renmin University of China. 

\bibliographystyle{ACM-Reference-Format}
\balance
\bibliography{icmr2022}


\end{document}